\documentclass[journal]{IEEEtran}

\IEEEoverridecommandlockouts

\usepackage{graphicx}
\usepackage{amsmath}
\usepackage{amssymb}
\usepackage{amsthm}
\usepackage{mathtools}
\usepackage{bm}
\usepackage{array}
\usepackage{booktabs}
\usepackage{tabularx}
\usepackage{multirow}
\usepackage{colortbl}
\usepackage{algorithm}
\usepackage{algorithmic}
\usepackage{cite}
\usepackage{url}
\usepackage{xcolor}
\usepackage[hidelinks]{hyperref}
\usepackage{makecell}

\graphicspath{{./}{figures/}}

\newtheorem{proposition}{Proposition}
\newtheorem{assumption}{Assumption}

\title{Push-and-Reach: Penetration-Bounded\\Dual-Arm Planning in Cluttered Canopies}

\author{Ege Gursoy$^{1,2}$ Robin Passama$^{1}$ Andrea Cherubini$^{1,3}$%

\thanks{This work was supported by ANR (PEPR PC2 PI2-IMM Organic Robotics)
under grant 22-EXOD-0003 and by the Défi Clé Robotique centrée sur
l’Humain funded by Région Occitanie, France.}%

\thanks{$^{1}$ LIRMM, Univ. Montpellier, CNRS, Montpellier, France.}%
\thanks{$^{2}$ LAAS-CNRS, Toulouse, France.}%
\thanks{$^{3}$ LS2N, Ecole Centrale de Nantes, CNRS, Nantes, France.}%
}

\begin{document}
\bstctlcite{IEEEexample:BSTcontrol}

\maketitle
\thispagestyle{empty}
\pagestyle{empty}

\begin{abstract}
Harvesting fruit in dense canopies presents challenges for both vision and manipulation. Even after a target is successfully located, foliage often obstructs the path to the fruit. Pushing through foliage without limit, however, risks damaging the plant or robot. We present \emph{Push-and-Reach}, a dual-arm planning method that turns a failed reach into a controlled intervention. When a reach trajectory fails collision-free verification, the detected contact points are clustered and ranked according to their influence on the planned reach. The second arm then pushes this obstruction under a hard constraint that bounds its penetration into the reconstructed canopy, and holds the foliage aside during the final reach. After the push, the canopy map is updated, the reach is replanned, and the final motion is executed after strict collision-free verification. On a real dual-arm robot, our method increases reach success from 33.3\% with collision-free planning alone to 86.7\% with a single bounded push and 93.3\% with up to three pushes, while maintaining measured penetration below the planned bound. In simulated canopy scenes, using the collision information from failed reaches leads to more feasible pushes than selecting obstructions based on cluster size or proximity. These results demonstrate the potential of dual-arm planning with bounded physical interaction to improve fruit accessibility in dense canopies.
\end{abstract}

\section{Introduction}
\label{sec:intro}

Although robotic fruit harvesting has matured on the perception side, access in dense canopies still remains a bottleneck. Dual-arm platforms~\cite{yoshida2022dualarm,zhang2024unifiedapple,lammers2024dualarm} promise higher throughput, although, the second arm generally stays idle during the harvesting reach, thus the canopy-access problem carries over from the single-arm setting. In a harvesting scenario, the planner faces two contact regimes at once, the reach itself should be verifiable as collision-free to limit any damage to the canopy or robot. Yet, clutter that blocks that reach may be compliant. Currently, strict-avoidance trajectory optimization does not return a solution when the corridor is fouled by foliage, and formulations that model the contact explicitly~\cite{posa2014direct,onol2020tuningfree,nemlekar2021lime} require contact-dynamics or permeability models, which are hard to fit to leaves and thin branches.

\begin{figure}[t]
    \centering
    \includegraphics[width=0.95\columnwidth]{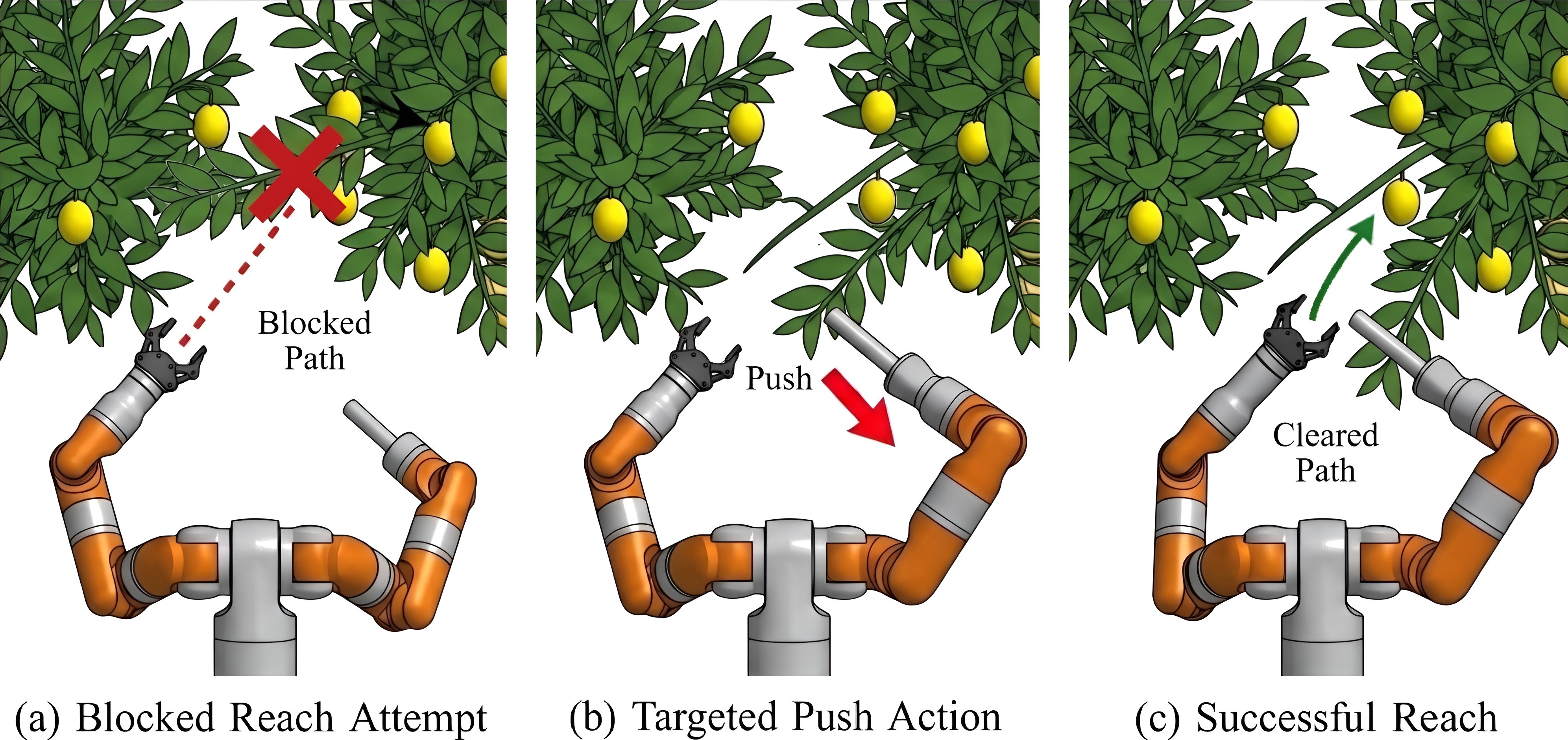}
    \vspace{-0.2cm}
    \caption{\emph{Push-and-Reach} in a cluttered canopy.}
    \vspace{-0.4cm}
    \label{fig:teaser}
\end{figure}

We imagine a pragmatic middle ground that does not require a stiffness model. When a strict reach attempt is rejected, the verifier algorithm returns where the trajectory failed. That information is structured enough, to drive a controlled intervention by the second arm. A push that displaces the dominant cluster of contacts can clear the corridor, and verifying the subsequent reach against the updated map keeps the executed final motion collision-free with respect to that map (Fig.~\ref{fig:teaser}). Following this idea, the reacher plans a soft-cost trajectory, and a verifier returns a collision locus along that trajectory if the motion is not clear. Density-based clustering~\cite{ester1996dbscan} groups the contacts, and an importance rule, which combines penetration depth, corridor proximity, temporal span, and joint-level sensitivity, picks one cluster as the push target. The pusher then plans a short push subject to a hard constraint that bounds its penetration into the mapped canopy by a prescribed depth $\delta$. While the pusher holds the foliage aside, the map is updated from the post-push depth frames, the reach is re-planned, and only the verified trajectory is executed.

In this work, we address the post-mapping segment of the pipeline. The target pose is therefore known, provided by an upstream perception stage, where the canopy is represented by an occupancy grid and a derived signed-distance field. We require the harvesting tool to approach the fruit without contact, as the contact can move the fruit away from its measured pose. A separate pusher can choose its contact pose from the obstruction geometry alone and hold the foliage during the reach, tools that tolerate leaf contact can instead reach through foliage~\cite{nemlekar2021lime}. We ask whether allowing the second arm to make a bounded contact with the foliage lets the reacher pass strict collision-free verification more often than collision-free planning alone. We validate the method on a dual-arm robot with a mock-up tree, over 6 fruit positions and 30 trials, and compare push-selection rules on additional simulated canopy scenes.

Our contributions are:
\begin{itemize}
    \item A dual-arm planning method that uses contact points, penetration depths, and surface normals from failed reach attempts to identify obstructing foliage and parameterize corrective push motions.

    \item A push planner with hard penetration constraints enforced during planning and at control rate, with a theoretical bound on executed penetration into the reconstructed canopy under assumptions.

    \item Experimental validation on a dual-arm robot in a mock-up tree, with ablation studies and comparisons of push-selection methods across simulated canopy scenes.
\end{itemize}

\section{Related Work}
\label{sec:relwork}

Perception and reconstruction have made target localization in cluttered canopies tractable, detection and segmentation pipelines extract fruit even from partial views~\cite{sa2016deepfruits,bargoti2017imageseg,koirala2019mangoyolo,santos2020grape}, and dense reconstruction, panoptic mapping, shape completion, and active perception accumulate evidence across viewpoints~\cite{pan2023panoptic,magistri2022contrastive,marangoz2022case,magistri2024icra,menon2023nbvsc,li2025enhancedviewplanning,chu2023alacs}. Such methods yield a target pose, but leave the access problem open, since the fruit can be localized while the robot still has no clear corridor.

Mapping the localized target to robot motion is the role of optimization-based planners such as CHOMP, STOMP, TrajOpt, and GPMP2, which trade smoothness, goal accuracy, and a collision cost defined over occupancy or signed-distance fields~\cite{zucker2013chomp,kalakrishnan2011stomp,schulman2014trajopt,dong2016gpmp2}, and the supporting infrastructure has matured through Octomap, iSDF, Voxfield, and nvblox~\cite{hornung2013octomap,ortiz2022isdf,pan2022voxfield,millane2024nvblox}. Collision avoidance has also been formulated through hard signed-distance constraints in model predictive control \cite{haffemayer2024model}. Strict planning over leaves and thin branches typically returns no solution, while a soft pass returns a motion that the strict verifier later rejects. Information about where trajectory failure is rarely propagated forward into a second action.

Contact-implicit optimization represents interaction directly, letting the optimizer decide when and where contact occurs through complementarity constraints~\cite{posa2014direct,onol2020tuningfree}, but requires a contact-dynamics model. Permeable-obstacle planning assigns a cost to passing through leaves in an RRT-based planner~\cite{nemlekar2021lime} and shares with our setting the broad observation that limited intrusion can be useful. For us, the pusher is allowed to penetrate up to a planned bound, while the reacher is verified against the post-push map, keeping the final reach auditable.

Physical interaction with the environment can also facilitate access to obstructed targets. The use of environmental contacts to manipulate deformable objects has been studied in \cite{zhu2019robotic}. In harvesting, movement-primitive approaches push occluding objects during picking~\cite{mghames2020ipromp}, learning-based methods displace foliage to reveal hidden fruit~\cite{zhang2023pushpastgreen,subedi2025findfruit}, and safe leaf manipulation has been studied with dedicated plant-interaction models and simulators~\cite{yao2025safeleaf,deng2024gazebo}. Other works explored vision-based dual-arm coordination for fruit harvesting~\cite{gursoy2023case} and used single-arm pushing to improve detection under occlusion~\cite{gursoy2024occlusionpush}. Point-cloud-based visual planning and servoing have been applied to vine pruning~\cite{gebrayel2026point}, and whole-system studies report success rates and ablate components under real-robot constraints~\cite{yoshida2022dualarm,zhang2024unifiedapple,lammers2024dualarm,magistri2024cluttered}. These works demonstrate the benefits of controlled interaction with foliage, but leave open the question of how to identify the foliage responsible for blocking the reach and displace it while limiting penetration into the canopy. \textit{Push-and-Reach} builds on these works by using collision information as a basis for planning physical interactions that make collision-free motions possible.

\section{Problem Statement}
\label{sec:problem}

\textit{Robot task}: The workspace $\mathcal{W}\subset\mathbb{R}^3$ contains a target fruit pose $\mathbf{T}^\star \in SE(3)$ expressed in a fixed base frame $B$. Two $n$-DoF manipulators operate in this workspace (here $n=7$). The \emph{reacher} $\mathcal{A}_r$ is responsible for the final approach to the fruit and the \emph{pusher} $\mathcal{A}_p$ displaces foliage when the reach is blocked. For each arm $a\in\{r,p\}$, the joint configuration is $\mathbf{q}_a\in\mathbb{R}^n$, and the combined configuration $\mathbf{q}=[\mathbf{q}_r^\top\ \mathbf{q}_p^\top]^\top\in\mathbb{R}^{2n}$ stacks both. Forward kinematics map $\mathbf{q}_a$ to the end-effector pose $\mathbf{T}_a(\mathbf{q}_a)\in SE(3)$. The reacher must bring $\mathbf{T}_r(\mathbf{q}_{r,T})$ to $\mathbf{T}^\star$ within tolerances $(\varepsilon_p,\varepsilon_R)$, satisfy joint and velocity limits, and pass a collision-free verification test against the post-interaction map. The pusher is allowed to penetrate the mapped canopy up to a planned bound to reconfigure compliant clutter. The two arms must also avoid each other, with $\min_{\ell\in\mathcal{A}_r,\ell'\in\mathcal{A}_p}\mathrm{dist}(\ell,\ell')\ge \delta_{\mathrm{arms}}$ at all times.

\textit{Environment representation}: Occupancy is a function $\Omega:\mathcal{W}\rightarrow[0,1]$ with threshold $\theta\in(0,1)$, the occupied set is $\mathcal{O}=\{\mathbf{x}\in\mathcal{W}\mid \Omega(\mathbf{x}) \ge \theta\}$. A signed-distance field (SDF) $d:\mathcal{W}\to\mathbb{R}$ is computed from $\Omega$ and $\theta$, with $d>0$ in free space and $d<0$ inside obstacles. We define the \emph{reach corridor} $\mathcal{R}\subset\mathcal{W}$ as the set of points within a fixed radius $r_{\mathcal{R}}$ of the nominal straight-line segment that joins the reacher end-effector start position to the target position of $\mathbf{T}^\star$ which approximates the volume the arm must traverse. For each link-obstacle pair $p$ of arm $a$ at configuration $\mathbf{q}_{a,t}$, the link-level signed distance is denoted $d_p(\mathbf{q}_{a,t})$. Let $\mathcal{P}$ denote the set of link-obstacle pairs evaluated by the distance query, where each pair $p\in\mathcal{P}$ associates a robot link with its nearest occupied region in the map. For arm $a$ at configuration $\mathbf{q}_{a,t}$, the signed distance of pair $p$ is denoted by $d_p(\mathbf{q}_{a,t})$. 

We build the occupancy map from depth images using ray casting to distinguish free, occupied, and unknown space. Unknown regions are treated as obstacles during planning and verification. After each push, we reconstruct the canopy map from new depth images while the pushing arm holds the foliage in place. We exclude the robot from the reconstruction and store fixed obstacles separately. The SDF is then computed from the occupied voxels and interpolated during trajectory optimization. Collision verification uses the voxel boundaries with finer spatial and temporal sampling. Therefore, our proposed method relies on the following assumptions:

\begin{assumption}[]\label{ass:first}
The foliage can be locally displaced by short, low-speed pushes, and the post-push map update reflects the changed geometry with enough fidelity for the reacher to replan against it, i.e. voxels cleared by the push are correctly identified as free space, within the limits of sensor accuracy.
\end{assumption}

\begin{assumption}[]\label{ass:err}
For a fixed reconstructed map, the SDF query error is bounded by $\varepsilon_d$ and the tracking error of every point on the pusher links and tool by $\varepsilon_t$. Every point on the reference geometry lies within $\rho_B$ of a spatial sample and moves at a maximum speed $v_B$. The trajectory is checked at both endpoints and at intervals of at most $h$, with the nonlinear distance constraint satisfied within a tolerance $\varepsilon_c$ at all the sampled points.
\end{assumption}

\section{Method}
\label{sec:method}

Algorithm~\ref{alg:pushreach} summarizes the pipeline which is detailed as a soft reach attempt (Sec.~\ref{sec:reacher}), whose rejection identifies the obstruction (Sec.~\ref{sec:cluster}), a penetration-bounded push (Sec.~\ref{sec:push}), and the coordination (Sec.~\ref{sec:coord}) and execution (Sec.~\ref{sec:execution}) of both arms.

\subsection{Reacher trajectory optimization}
\label{sec:reacher}

The reacher first plans a trajectory toward the fruit using soft collision costs. If the trajectory fails collision-free verification, the detected contacts identify the foliage obstructing the reach. The reach trajectory is discretized as $\tau_r=\{\mathbf{q}_{r,0},\dots,\mathbf{q}_{r,T}\}$ over a fixed horizon $T$ and obtained by minimizing the following cost function:
\begin{align}
\min_{\tau_r}\;
& J_r(\tau_r) = w_s J_{\mathrm{s}}(\tau_r)
+ w_x J_{\mathrm{goal}}(\tau_r)
+ w_c J_{\mathrm{coll}}(\tau_r, s_r) \label{eq:reach_obj}\\
\mathrm{s.t.}\;
& \mathbf{q}_{r,\min} \le \mathbf{q}_{r,t} \le \mathbf{q}_{r,\max}, \quad t=0,\dots,T \nonumber\\
& |q^{j}_{r,t}-q^{j}_{r,t-1}| \le \Delta_{\max}, \quad j=1,\dots,n,\ t=1,\dots,T \nonumber\\
& d_{rp}(\mathbf{q}_{r,t},\mathbf{q}_{p,t}) \ge \delta_{\mathrm{arms}},\quad t=0,\dots,T \nonumber
\end{align}
The smoothness term $J_{\mathrm{s}}(\tau_r) = \sum_{t=1}^{T} \|\mathbf{q}_{r,t}-\mathbf{q}_{r,t-1}\|^2$ penalizes large joint changes. The goal term $J_{\mathrm{goal}}$ penalizes the terminal-pose error, the squared norm of the $SE(3)$ logarithm of $T_r(\mathbf{q}_{r,T})^{-1}T^\star$. The collision term is the hinge
\begin{equation}
\begin{aligned}
J_{\mathrm{coll}}(\tau_r,s_r) &= \sum_{t=0}^{T}\sum_{p\in\mathcal{P}}
\phi_{s_r}\!\left(d_p(\mathbf{q}_{r,t})\right)\\
\phi_{s}(d) &= \big[\max(0,\, s-d)\big]^2
\end{aligned}
\label{eq:hinge}
\end{equation}
where $\phi_s$ penalizes signed distances below the margin $s$ quadratically, with a  buffer to smooth the transition at $d=s$. The constraints enforce joint position and velocity limits through the maximum joint displacement $\Delta_{\max}$ between consecutive waypoints, and maintain a minimum separation between the two arms, with $d_{rp}$ denoting the minimum distance between their links. Eq.~\eqref{eq:reach_obj} is solved using a gradient-based trajectory optimizer with line search~\cite{schulman2014trajopt}, initialized with the previous reach trajectory.

\begin{algorithm}[t]
\caption{\textit{Push-and-Reach}}
\label{alg:pushreach}
\footnotesize
\begin{algorithmic}[1]
\REQUIRE Initial map $\Omega_0$, target pose $\mathbf{T}^\star$, maximum pushes $K_{\max}$
\ENSURE Reach trajectory, push sequence, success flag
\STATE $\Omega\leftarrow\Omega_0$, $k\leftarrow0$
\WHILE{true}
    \STATE Plan $\tau_r$ using \eqref{eq:reach_obj}
    \STATE $(\textit{clear}, \mathcal{C})\leftarrow\mathrm{StrictCheck}(\tau_r,\Omega)$
    \IF{\textit{clear}}
        \STATE Execute $\tau_r$ and retreat the pusher if holding
        \RETURN $(\tau_r,\{\tau_{p,j}\}_{j=1}^{k}, \textit{success})$
    \ENDIF
    \STATE \textbf{if} $k=K_{\max}$ \textbf{then break}
    \IF{pusher is holding foliage}
        \STATE Release foliage, retreat, and update $\Omega$
        \STATE \textbf{continue}
    \ENDIF
    \STATE Cluster contacts in $\mathcal{C}$ using DBSCAN
    \STATE Select the eligible cluster with highest $I_k$ using \eqref{eq:imp}
    \STATE Plan $\tau_{p,k+1}$ using \eqref{eq:push_obj}
    \STATE \textbf{if} no feasible push \textbf{then break}
    \STATE Execute push and hold foliage
    \STATE Update $\Omega$ and set $k\leftarrow k+1$
\ENDWHILE
\STATE Release and retreat the pusher if holding
\RETURN $(\emptyset,\{\tau_{p,j}\}_{j=1}^{k}, \textit{failure})$
\end{algorithmic}
\end{algorithm}

\begin{figure*}[t]
    \centering
    \includegraphics[width=0.99\textwidth
    ,trim={0 0 0 0.8cm},
    clip]{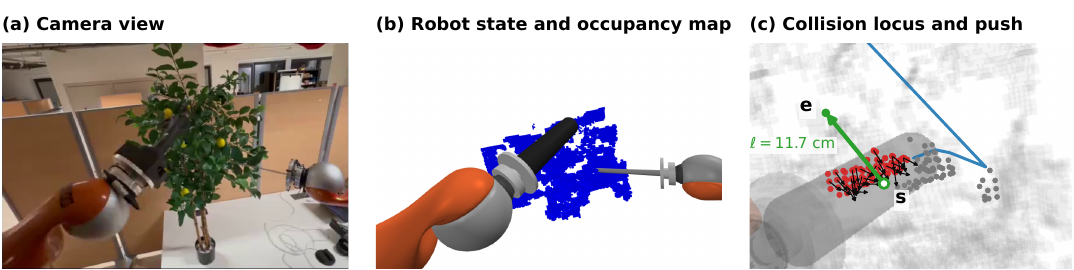}
    \caption{Collision-locus extraction and bounded-push synthesis towards target fruit $T^\star$ on robot data (left: camera view, middle: robot state and occupancy map, right: close-up). On right, the failed reach trajectory is shown in blue, verifier contacts are clustered into the dominant obstruction $k^\star$ in red and other contacts in gray. Black arrows show the per-contact surface normals, the green arrow is the push from $\mathbf{s}_{k^\star}$ to $\mathbf{e}_{k^\star}$ along $\hat{\mathbf{d}}_{k^\star}=-\bar{\mathbf{n}}_{k^\star}$, into the foliage, away from the corridor.}
    \label{fig:cluster}
\end{figure*}

\subsection{Obstruction Identification}
\label{sec:cluster}

The collision verifier evaluates $\tau_r$ using more accurate signed-distance queries and finer trajectory sampling than the optimizer.

A strict collision check along a candidate trajectory returns
\begin{equation}
\mathcal{C}=\{(\mathbf{p}_i,\mathbf{n}_i,d_i,t_i)\}_{i=1}^{N}
\end{equation}
where $i=1,\dots,N$ are detected (near-)contacts, $\mathbf{p}_i\in\mathbb{R}^3$ is the obstacle-side closest point, $\mathbf{n}_i\in\mathbb{S}^2$ the outward surface normal, $d_i$ the signed distance, and $t_i\in\{0,\dots,T\}$ the index of the trajectory waypoint $\mathbf{q}_{r,t_i}$ at which contact $i$ was detected (several contacts may share a waypoint).

From the contact set $\mathcal{C}$, the contact points form the collision locus $\mathcal{L}=\{\mathbf{p}_i\}_{i=1}^{N}$. Nearby contacts typically correspond to leaves, branches, or regions of dense foliage. DBSCAN~\cite{ester1996dbscan} groups these contacts into clusters $\{\mathcal{K}_k\}$ and discards isolated points (Fig.~\ref{fig:cluster}), where $\mathcal{K}_k$ contains the contact indices belonging to cluster $k$.

The signed-distance gradient with respect to joints follows the standard kinematic expression
\begin{equation}
\frac{\partial d_p}{\partial \mathbf{q}_a}(\mathbf{q}_{a,t})
= \mathbf{n}_{p,t}^\top J_{p}(\mathbf{q}_{a,t})
\label{eq:jacgrad}
\end{equation}
where $J_{p}\in\mathbb{R}^{3\times n}$ is the translational Jacobian of the link in pair $p$, evaluated at the link point closest to the obstacle, and $\mathbf{n}_{p,t}$ is the corresponding outward obstacle normal. The norm of \eqref{eq:jacgrad} measures how strongly joint motion changes the contact distance.

For each contact $i$, the sensitivity $g_i$ of its signed distance to joint motion and its penetration depth $e_i$ are given by
\begin{equation}
g_i = \left\|
\frac{\partial d_{p(i)}}{\partial \mathbf{q}_r}
(\mathbf{q}_{r,t_i})
\right\|,
\qquad
e_i = \max(0,-d_i)
\end{equation}
where $p(i)$ denotes the collision pair associated with contact $i$. The contact weight is defined as
\begin{equation}
w_i = g_i + e_i + 0.5\,\frac{t_i}{T}
\label{eq:weights}
\end{equation}
giving higher weights to contacts that are sensitive to joint motion, have deeper penetrations, or occur toward the end of the reach. The centroid $\bar{\mathbf{p}}_k$ and weighted surface normal $\bar{\mathbf{n}}_k$ of each cluster are computed as
\begin{equation}
\bar{\mathbf{p}}_k =
\frac{1}{|\mathcal{K}_k|}
\sum_{i\in\mathcal{K}_k}\mathbf{p}_i,
\quad
\bar{\mathbf{n}}_k =
\frac{\sum_{i\in\mathcal{K}_k}w_i\mathbf{n}_i}
{\left\|\sum_{i\in\mathcal{K}_k}w_i\mathbf{n}_i\right\|}
\end{equation}
where $|\mathcal{K}_k|$ is the number of contacts in the cluster. To determine which cluster to push, an importance score combines the mean sensitivity $\bar{g}_k$, mean penetration depth $\bar{e}_k$, proximity to the reach corridor $\rho_k$, and temporal span $\kappa_k$,
\begin{equation}
I_k = 0.10\,\bar{g}_k
+ 0.10\,\bar{e}_k
+ 0.015\,\rho_k
+ 0.05\,\kappa_k
\label{eq:imp}
\end{equation}
with
\begin{equation}
\begin{aligned}
\rho_k &=
\left(1+
\frac{\mathrm{dist}(\bar{\mathbf{p}}_k,\mathcal{R})}
{r_{\mathcal{R}}}\right)^{-1}\\
\kappa_k &=
\frac{\max_{i\in\mathcal{K}_k}t_i-
\min_{i\in\mathcal{K}_k}t_i}{T}
\end{aligned}
\end{equation}
The proximity term favors obstructions near the reach corridor, while the temporal span favors obstructions encountered over a larger portion of the trajectory. Combining the four terms accounts for both the location of the foliage and its effect on the reacher's motion.

Since the surface normals are computed from the reconstructed canopy, their directions may vary within a cluster. Their directional agreement is measured by
\begin{equation}
\gamma_k =
\frac{\left\|\sum_{i\in\mathcal{K}_k}w_i\mathbf{n}_i\right\|}
{\sum_{i\in\mathcal{K}_k}w_i}
\end{equation}
where $\gamma_k=1$ indicates that all positively weighted normals point in the same direction. Clusters with zero total weight or $\gamma_k<\gamma_{\min}$ are excluded from selection to avoid poorly defined push directions. The cluster $k^\star$ with the highest importance score among the remaining clusters is selected as the push target.

\subsection{Push synthesis with bounded penetration}
\label{sec:push}

The weighted surface normal $\bar{\mathbf{n}}_{k^\star}$ points outward from the foliage toward the obstructed reacher link. The push direction is therefore defined as $\hat{\mathbf{d}}_{k^\star}=-\bar{\mathbf{n}}_{k^\star}$ to displace the foliage away from the reach corridor. To first order, displacing a contact point by $\Delta\hat{\mathbf{d}}_{k^\star}$, with $\Delta>0$, increases its distance to the reacher by $\Delta\,\mathbf{n}_i^\top\bar{\mathbf{n}}_{k^\star}$ when the two normals point in similar directions.

The push is defined as a straight segment of length $\ell_{k^\star}$ between a start position $\mathbf{s}_{k^\star}$ and an end position $\mathbf{e}_{k^\star}$. The start position is placed at a distance $\delta_s>0$ from the cluster centroid on the side facing the reach corridor,
\begin{equation}
\mathbf{s}_{k^\star} = \bar{\mathbf{p}}_{k^\star} - \delta_s \hat{\mathbf{d}}_{k^\star},
\qquad
\mathbf{e}_{k^\star} = \mathbf{s}_{k^\star} + \ell_{k^\star}\hat{\mathbf{d}}_{k^\star}
\label{eq:seg}
\end{equation}
The push length is determined by the importance score $I_{k^\star}$, expressed in meters, and limited by the maximum push length $\ell_{\max}$,
\begin{equation}
\ell_{k^\star} = \min\{\ell_{\max}, I_{k^\star}\}
\label{eq:len}
\end{equation}
Clusters with higher importance receive longer pushes. The push length includes the initial distance $\delta_s$ to the foliage and is independent of the maximum allowed penetration $\delta>0$. Pushes with $\ell_{k^\star}\le\delta_s$ are rejected. The end-effector orientation $R_{k^\star}$ aligns the tool $x$-axis with the push direction $\hat{\mathbf{d}}_{k^\star}$, with the roll determined by projecting the world vertical onto the plane perpendicular to the push direction.

The pusher trajectory $\tau_p$ consists of an approach to $\mathbf{s}_{k^\star}$, a straight push to $\mathbf{e}_{k^\star}$, and a hold at $\mathbf{e}_{k^\star}$ until the reacher completes its motion. The trajectory is obtained by solving
\begin{align}
\min_{\tau_p}\; & J_p(\tau_p) = w_s J_{\mathrm{s}}(\tau_p) + w_x J_{\mathrm{t}}(\tau_p) + w_c J_{\mathrm{coll}}(\tau_p, s_p) \label{eq:push_obj}\\
\mathrm{s.t.}\;&\text{joint and arm separation constraints as in \eqref{eq:reach_obj}}\nonumber\\
& d_p(\mathbf{q}_{p,t}) \ge -\delta,\quad p\in\mathcal{P},\ t=0,\dots,T \nonumber
\end{align}
where the smoothness term $J_{\mathrm{s}}$ and the cost weights follow \eqref{eq:reach_obj}, and $J_{\mathrm{t}}$ penalizes Cartesian deviation from the push segment. The collision cost $J_{\mathrm{coll}}$ uses the hinge function \eqref{eq:hinge} with a negative margin $s_p=-\delta$. The hard constraint limits the penetration of pusher links into the canopy to $\delta$.

With $s_p=-\delta$, the collision cost is zero for all distances satisfying $d\ge-\delta$, allowing the optimizer to plan contact with the foliage within the penetration bound. With $s_p=0$, any penetration incurs a cost, discouraging the contact needed to displace the foliage. 

\subsection{Iterative coordination}
\label{sec:coord}

After each push, the pusher holds the foliage in place while the canopy map is updated and a new reach trajectory is planned. If the new trajectory passes collision-free verification, the reacher executes it and the pusher then releases the foliage and retreats. Otherwise, the pusher releases the foliage, and the map is updated again to account for any foliage motion after release. The next push is planned using the collision information from a new reach attempt against the updated map. The process continues until a collision-free reach is found or the maximum number of pushes $K_{\max}$ is reached. A single push ($K_{\max}=1$) is used by default to minimize physical interaction with the canopy. Later on, up to three pushes ($K_{\max}=3$) are used to determine whether repeated interactions improve access to more obstructed targets.
\subsection{Execution and penetration bound}
\label{sec:execution}

Both arms track their planned trajectories using a single joint-velocity QP. The trajectories $\tau_r$ and $\tau_p$ are time-parameterized to obtain the joint position and velocity references $\mathbf{q}_{\mathrm{ref}}(t)$ and $\dot{\mathbf{q}}_{\mathrm{ref}}(t)$. At each control step $\Delta t$, the commanded joint velocities are computed by solving
\begin{align}
\min_{\dot{\mathbf{q}}}\ & \big\|\dot{\mathbf{q}} - \left(\dot{\mathbf{q}}_{\mathrm{ref}} + K_p(\mathbf{q}_{\mathrm{ref}}-\mathbf{q})\right)\big\|^2 \label{eq:qp}\\
\mathrm{s.t.}\;
& \mathbf{q}_{\min}\le\mathbf{q}+\dot{\mathbf{q}}\Delta t\le\mathbf{q}_{\max}\nonumber\\
& \dot{\mathbf{q}}_{\min}\le\dot{\mathbf{q}}\le\dot{\mathbf{q}}_{\max}\nonumber\\
& d_{rp}(\mathbf{q}) + \nabla d_{rp}(\mathbf{q})^\top\dot{\mathbf{q}}\,\Delta t \ge \delta_{\mathrm{arms}}\nonumber\\
& d_p(\mathbf{q}) + \nabla d_p(\mathbf{q})^\top\dot{\mathbf{q}}\,\Delta t \ge -\delta \quad \text{(pusher links)}\nonumber
\end{align}
where $K_p$ is the proportional tracking gain and $\mathbf{q}$ is the measured joint configuration. The penetration constraint enforced during planning and at control rate gives the following bound on the executed motion under the stated error and sampling assumptions.

\begin{proposition}\label{prop:pen}
Let $d_p^\Omega$ denote the minimum exact signed distance between the pusher link associated with pair $p$ and the fixed reconstructed map, and $\mathbf{q}_p^{\mathrm{exec}}(t)$ the executed joint configuration of the pusher. Under Assumption~\ref{ass:err}, the maximum executed penetration satisfies
\begin{equation}
\begin{aligned}
&\max_{t,p}\max\bigl\{0,-d_p^\Omega(\mathbf{q}_p^{\mathrm{exec}}(t))\bigr\}\\
&\qquad\le \delta+\varepsilon_c+\varepsilon_t+\varepsilon_d
+\rho_B+\frac{v_Bh}{2}
\end{aligned}
\label{eq:penbound}
\end{equation}
\end{proposition}

\begin{proof}
At the sampled body points and trajectory times, the exact signed distance is bounded below by $-\delta-\varepsilon_c-\varepsilon_d$. Accounting for tracking error, spatial sampling, and motion between consecutive checks, any point on the pusher during execution is at most $\varepsilon_t+\rho_B+v_Bh/2$ away from a reference sample. Since the signed distance cannot change by more than the spatial displacement, the maximum penetration is bounded by \eqref{eq:penbound}.
\end{proof}
\section{Experiments}
\label{sec:exp}

We validate \textit{Push-and-Reach} on a dual-arm robot in a cluttered canopy. The experiments are designed to show that a single bounded push raises verified reach success beyond collision-free planning alone and that the executed penetration respects the planned bound. An ablation study evaluates the contribution of each component, followed by a comparison of different cluster selection methods and an evaluation of robustness in simulated canopy scenes.

Video demonstrations of the experiments are available at: \url{https://peertube.laas.fr/w/8zaLhwr5PmswVLNEZ6fpXi}.

\subsection{Experimental setup and evaluation}
\label{sec:setup}
\label{sec:protocol}

The experimental platform consists of two 7-DoF KUKA LWR4 arms. The pusher carries a rigid metal rod, and the reacher carries a cylindrical end-effector representing a harvesting tool. A fixed Intel RealSense D435 RGB-D camera provides depth images to reconstruct the canopy within a $3\!\times\!3\!\times\!3~\mathrm{m}^3$ workspace at a voxel resolution of 2~cm. Collision verification uses 4$\times$ finer sampling and stricter thresholds than trajectory optimization. After each push, the canopy map is reconstructed from ten depth frames acquired at 30~Hz.

The mock-up tree contains compliant foliage and six fruits at different positions. We perform five trials per target, for a total of 30 trials per method, starting from a home configuration with a small random perturbation ($\pm 2$~cm). Figure~\ref{fig:sequence} shows an example of the method executed on the robot.

The planning and execution parameters are given in the appendix (Section~\ref{app:params}). The penetration bound is $\delta=5$~cm. Both arms use the same trajectory cost weights, with a positive collision margin for the reacher and a negative margin for the pusher to allow contact within the penetration bound.

We evaluate Push-and-Reach with a single push ($K_{\max}=1$) and separately with up to three pushes ($K_{\max}=3$) to assess the benefit of repeated interactions with the canopy. 

A trial is successful if the reach trajectory passes collision-free verification and the robot reaches the target within the position and orientation tolerances $(\varepsilon_p,\varepsilon_R)$. Both planning and execution failures are counted as unsuccessful trials.

Clearance is measured as the minimum signed distance between the reacher and the canopy along the executed trajectory, using the map against which the reach was verified. Penetration is measured as the maximum depth of any pusher link or tool point into the pre-push map throughout the approach, push, hold, and retreat phases, and is compared with the bound in Proposition~\ref{prop:pen}.

To measure the reduction in foliage obstructing the reach corridor, we define the obstruction metric $\Phi$ and its relative reduction $\mathrm{Red}_\Phi$
\begin{equation}
\begin{aligned}
\Phi(\Omega) &= \sum_{\mathbf{v}\in\mathcal{R}}
\max(0,\Omega(\mathbf{v})-\theta)\\
\mathrm{Red}_\Phi &= 1-\frac{\Phi(\Omega_k)}{\Phi(\Omega_0)}
\end{aligned}
\end{equation}
where $\Omega_0$ and $\Omega_k$ are the canopy maps before interaction and after $k$ pushes. Both values of $\Phi$ are computed over the corridor voxels observed in both maps. We also report the total duration of each trial.

Success rates are reported with Wilson 95\% confidence intervals~\cite{wilson1927probable}. All methods use the same 30 initial configurations and reach planner, allowing direct comparison using a McNemar test~\cite{mcnemar1947note}, with a significance threshold of $p<0.05$.

\subsection{Single-push results}
\label{sec:results}
\label{sec:penetration}

We first evaluate Push-and-Reach with a maximum of one push per trial to determine whether a single interaction with the foliage is sufficient to reach obstructed targets. Table~\ref{tab:per_target} presents the results for each target in order of increasing corridor obstruction.

Allowing the second arm to push the foliage makes most of the initially blocked reaches possible, more than doubling the overall success rate compared with collision-free planning alone (33.3\% to 86.7\%). The benefit is particularly clear for targets T$_1$, T$_4$, and T$_3$, where most initial reaches fail verification. Most of the remaining failures occur at T$_6$ (Section~\ref{sec:multipush}).

\begin{table}[!ht]
\caption{Single-push results ordered by initial corridor obstruction.}
\label{tab:per_target}
\centering
\setlength{\tabcolsep}{2.0pt}
\footnotesize
\begin{tabular}{@{}cccccccc@{}}
\toprule
\textbf{ID} &
$\bm{\Phi(\Omega_0)}$ &
\textbf{Succ./Tri.} &
\textbf{Clear} &
\makecell[c]{\textbf{Push}\\\textbf{success}} &
\makecell[c]{\textbf{Push}\\\textbf{fail}} &
\makecell[c]{\textbf{Pre-}\\\textbf{push}} &
\textbf{Time}\,[s] \\
\midrule
T$_2$ & $0.12$ & 5/5 & 5 & 0 & 0 & 0 & $6.1\!\pm\!0.4$ \\
T$_5$ & $0.21$ & 5/5 & 4 & 1 & 0 & 0 & $12.0\!\pm\!4.7$ \\
T$_1$ & $0.38$ & 5/5 & 1 & 4 & 0 & 0 & $23.4\!\pm\!2.1$ \\
T$_4$ & $0.47$ & 4/5 & 0 & 4 & 0 & 1 & $22.5\!\pm\!3.0$ \\
T$_3$ & $0.55$ & 5/5 & 0 & 5 & 0 & 0 & $28.7\!\pm\!3.2$ \\
T$_6$ & $0.71$ & 2/5 & 0 & 2 & 3 & 0 & $35.2\!\pm\!4.1$ \\
\midrule
\textit{Overall} & $0.41$ & 26/30 ($86.7\%$) & 10 & 16 & 3 & 1 & $21.3\!\pm\!10.4$ \\
\bottomrule
\end{tabular}
\end{table}

\begin{figure}[t]
    \centering
    \scriptsize\renewcommand{\arraystretch}{0.9}\setlength{\tabcolsep}{0pt}
    \begin{tabular}{*{3}{>{\centering\arraybackslash}p{0.33\columnwidth}}}
    Initial & Push and hold & Reach\\
    0-16.5~s & 16.5-21.7~s & 21.7-27.5~s
    \end{tabular}\\[1pt]
    \includegraphics[width=0.99\columnwidth]{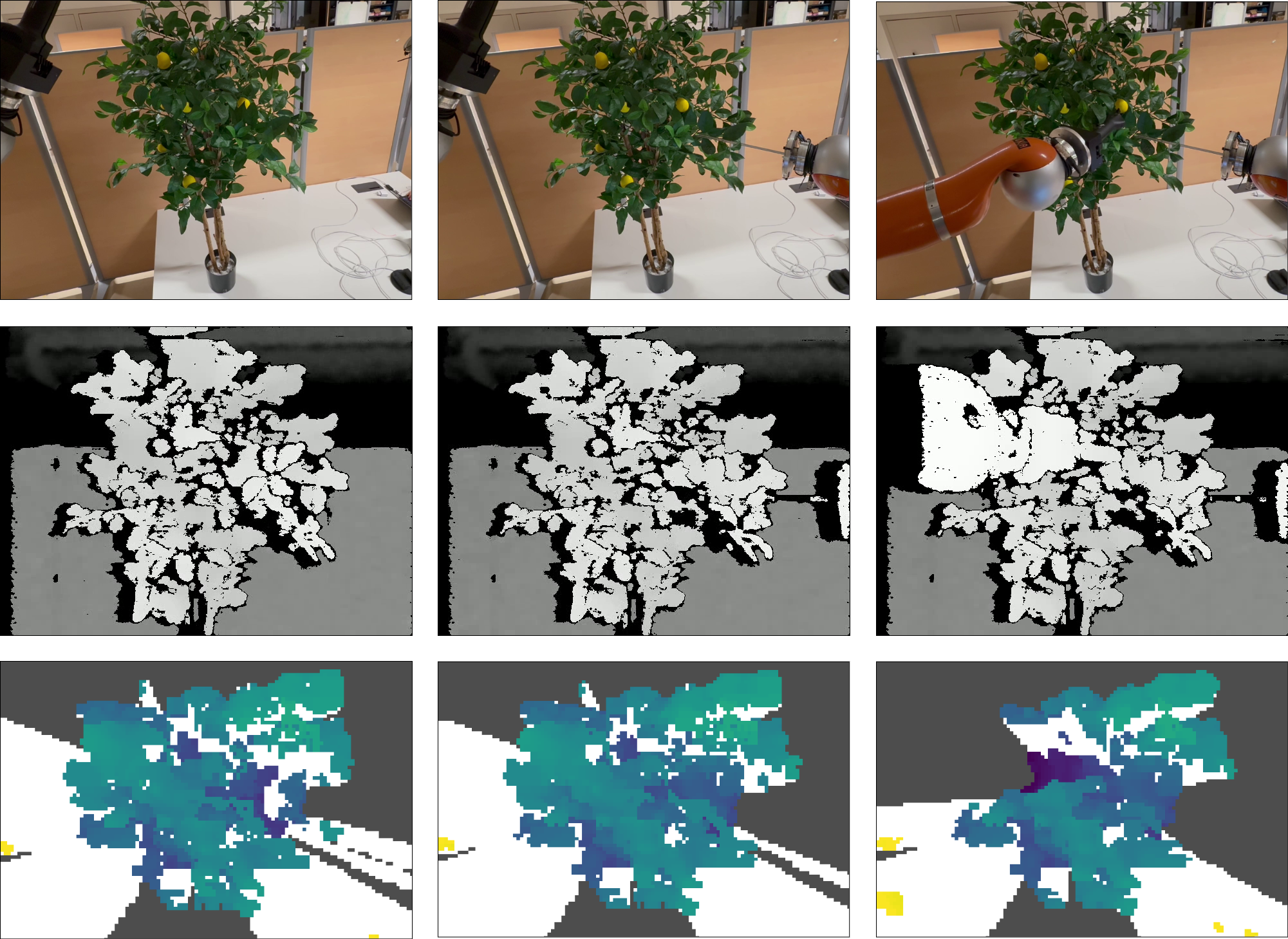}
    \caption{Real-world execution of \textit{Push-and-Reach} on a mock-up lemon tree (target T$_3$). Left to right: the initial reacher attempt, which fails strict verification, the pusher approach along $\hat{\mathbf{d}}_{k^\star}$, push execution, map update, and the verified reach to the fruit.}
    \label{fig:sequence}
\end{figure}

Figure~\ref{fig:sequence} illustrates a successful trial for target T$_3$. The initial reach fails verification, and the pusher displaces the selected obstruction, reducing the corridor obstruction by 54\%. While the pusher holds the foliage aside, the reacher plans and executes a collision-free trajectory against the updated canopy map.

To assess whether the planned penetration bound is respected during execution, we recompute the maximum penetration of each push against the pre-push map using the collision verifier with four times finer sampling than the optimizer. All 19 executed pushes remain below the prescribed penetration bound (4.7~cm maximum penetration for a bound of 5~cm), while all executed reaches maintain positive clearance from the reconstructed canopy.

The improvement in reach success is therefore achieved while respecting the prescribed geometric penetration limit, showing that limited physical interaction with the foliage is sufficient to clear the reach corridor in most trials.

\subsection{Baseline comparison and ablations}
\label{sec:ablation}

To evaluate the contribution of each component, we consider the following variants.
\begin{itemize}
    \item \textit{No-Push} performs collision-free reach planning without interacting with the foliage.
    \item \textit{Random-Cluster} selects a contact cluster at random.
    \item \textit{Count-Only} selects the cluster containing the largest number of contacts.
    \item \textit{No-Margin} uses a pusher collision margin of $s_p=0$ to penalize penetration while maintaining the hard penetration bound.
    \item \textit{No-Replan} verifies the original reach trajectory against the updated canopy map after the push without replanning.
\end{itemize}
Random-Cluster and Count-Only change only the cluster selection rule. The push length is still computed using \eqref{eq:imp} and \eqref{eq:len}.

The results are summarized in Table~\ref{tab:baselines}.

The largest improvement comes from allowing the second arm to interact with the foliage. The negative collision margin further improves reach success by allowing penetration within the hard bound without penalizing it. When a zero collision margin is used (\textit{No-Margin}), the pusher tends to stop before sufficiently displacing the foliage, even though the same penetration bound is enforced in both methods.

\begin{table}[!ht]
\caption{Baseline comparison and ablation across 30 trials.}
\label{tab:baselines}
\centering
\setlength{\tabcolsep}{3.0pt}
\footnotesize
\begin{tabular}{@{}lcccc@{}}
\toprule
\textbf{Method} &
\textbf{Success} &
\makecell[c]{\textbf{95\% CI}$[\%]$} &
\makecell[c]{$\bm{p}$ \textbf{vs Ours}} &
\makecell[c]{$\bm{-d_p}$\,[cm]} \\
\midrule
\textit{No-Push}        & $10/30\,(33.3\%)$ & $[19,51]$ & $<\!0.001$ & - \\
\textit{Random-Cluster} & $17/30\,(56.7\%)$ & $[39,73]$ & $0.004$ & $4.7$ \\
\textit{No-Margin}      & $19/30\,(63.3\%)$ & $[46,78]$ & $0.02$ & $0.4$ \\
\textit{Count-Only}     & $22/30\,(73.3\%)$ & $[56,86]$ & $0.13$ & $4.5$ \\
\textit{No-Replan}      & $23/30\,(76.7\%)$ & $[59,88]$ & $0.25$ & $4.4$ \\
\textit{Push-and-Reach} & $\mathbf{26/30\,(86.7\%)}$ & $[70,95]$ & - & $4.3$ \\
\bottomrule
\end{tabular}
\end{table}

The proposed importance score also significantly improves success compared with random cluster selection. On targets T$_4$ and T$_6$, the kinematic sensitivity term helps identify obstructions that have a greater influence on the reacher's motion, even when they contain fewer contact points. Both the proposed ranking and post-push replanning improve the observed success rate compared with \textit{Count-Only} and \textit{No-Replan}, respectively, although these differences are not statistically significant over the 30 trials.

The ablation study establishes the benefit of permitting contact within a prescribed penetration bound and selecting push targets based on the collision geometry of the attempted reach. The contribution of the individual terms in the importance score is further investigated in simulation across a larger number of canopy configurations.

\subsection{Repeated pushes and failure analysis}
\label{sec:multipush}
\label{sec:failure}

Single-push failures mainly occur when the pusher cannot approach the selected obstruction or when multiple layers of foliage block the reach. At T$_4$, the pusher cannot find a collision-free approach to the selected cluster, preventing the planned interaction.

To evaluate whether repeated pushes can improve access to targets surrounded by multiple layers of foliage, we repeat the experiments with up to three pushes per trial ($K_{\max}=3$).

The additional pushes improve the overall success rate (from 86.7\% to 93.3\%) by recovering two of the three unsuccessful trials at T$_6$, although the improvement is not statistically significant over the 30 trials. At T$_6$, the first push successfully displaces the outer layer of foliage, but a deeper layer continues to obstruct the reach corridor. Each push exposes a deeper layer of foliage, allowing the subsequent reach attempt to identify the remaining obstruction.

Allowing additional pushes does not resolve the inaccessible approach at T$_4$, and one trial at T$_6$ remains unsuccessful within the three-push budget. In all unsuccessful trials, the final reach trajectory fails collision-free verification and is not executed.

\subsection{Push selection on depth-reconstructed scenes}
\label{sec:selection}

\begin{figure}[!ht]
    \centering
    \includegraphics[width=0.99\columnwidth]{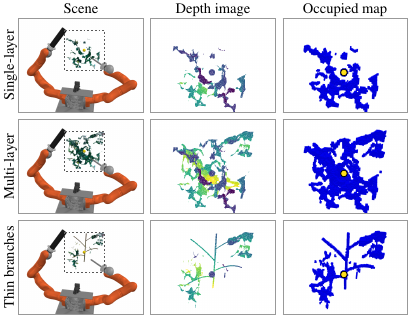}
    \caption{Simulated canopy scenes, depth images, and occupancy maps.}
    \label{fig:scenes}
\end{figure}

To evaluate the proposed push-selection rule across a wider range of canopy geometries than those available on the robot, we generate 600 scenes divided equally into three categories of foliage and branches (Fig.~\ref{fig:scenes}). Each category includes five occupancy levels, ranging from sparse to dense clutter. All scenes have a blocked initial reach and at least one eligible contact cluster. The canopy maps are reconstructed using ten simulated depth frames with 5~mm noise and 8\% missing pixels, following Section~\ref{sec:problem}. The planners use the reconstructed maps, while the true canopy geometry is used for independent verification.

We compare the proposed importance score with four cluster selection rules. \textit{Random} selects a cluster at random, \textit{Count-Only} selects the largest cluster, \textit{Deepest} selects the cluster with the largest mean penetration depth, and \textit{Corridor-Only} selects the cluster closest to the reach corridor. All methods use the same candidate pushes generated using \eqref{eq:seg} and \eqref{eq:len}. We also remove each term of \eqref{eq:imp} individually to assess its contribution to cluster selection.

Each selected push is evaluated according to three criteria.
\begin{itemize}
    \item \textit{Useful} if displacing the canopy voxels within 2~cm of the selected contacts by $(\ell_k-\delta_s)\hat{\mathbf{d}}_k$ reduces the total squared penetration along the rejected reach and new contacts.
    \item \textit{Depth-map feasible} if the approach is collision-free and the push satisfies all constraints on the reconstructed map.
    \item \textit{Reference-valid} if the same trajectory also satisfies these requirements on the true canopy geometry.
\end{itemize}

\begin{table}[!ht]
\caption{Push selection on depth-reconstructed scenes.}
\label{tab:selection}
\centering
\setlength{\tabcolsep}{0.0pt}
\footnotesize
\begin{tabular*}{\columnwidth}{@{\extracolsep{\fill}}lccc@{}}
\toprule
\textbf{Selector} &
\textbf{Useful} &
\makecell[c]{\textbf{Depth-map}\\\textbf{feas.}} &
\makecell[c]{\textbf{Ref.-valid}\\\textbf{95\% CI} $[\%]$} \\
\midrule
Random        & $245$ & $315$ & $298$ $[45.7,53.7]$ \\
Count-Only    & $482$ & $387$ & $366$ $[57.0,64.8]$ \\
Deepest       & $495$ & $400$ & $375$ $[58.6,66.3]$ \\
Corridor-Only & $518$ & $417$ & $391$ $[61.3,68.9]$ \\
Full rule     & $\mathbf{566}$ & $\mathbf{474}$ & $\mathbf{447}$ $[70.9,77.8]$ \\
\multicolumn{3}{@{}l}{\quad without sensitivity / depth / proximity / span} & $397$ / $437$ / $416$ / $443$ \\
\bottomrule
\end{tabular*}
\end{table}

The results in Table~\ref{tab:selection} show that the proposed importance score selects useful and feasible pushes more often than the individual selection rules. The largest reductions in performance occur when kinematic sensitivity or corridor proximity is removed from the importance score. Both terms help identify obstructions according to their influence on the planned reach. Combining them with penetration depth and temporal span improves the selection of valid pushes across different canopy geometries.

Reconstruction errors also affect push feasibility, particularly in scenes containing thin branches that are difficult to capture in depth images. A small proportion of pushes accepted on the reconstructed maps fail verification against the true geometry (5.7\%).

The sensitivity of the push direction to surface-normal errors is evaluated separately in the appendix (Section~\ref{app:normal-errors}).

\section{Conclusion}
\label{sec:conclusion}

We presented \textit{Push-and-Reach}, a dual-arm planning method that uses collision information from a failed reach to plan a corrective push with bounded penetration into the canopy. Experiments on a real dual-arm robot show that a single push more than doubles the reach success rate compared with collision-free planning alone, while keeping the measured penetration below the prescribed bound. The simulation results further show the benefit of selecting push targets based on their influence on the planned reach.

The current experiments are limited to an indoor mock-up tree. Field experiments are needed to evaluate the method under different foliage stiffnesses, lighting conditions, and environmental motion. The simulated scenes extend the evaluation to a wider range of canopy geometries but do not model foliage deformation during contact. More realistic interactions could be evaluated using plant simulation models~\cite{deng2024gazebo}. The penetration bound is geometric and defined with respect to the reconstructed canopy, making force and tactile sensing valuable additions for limiting plant damage.

Future work includes combining repeated pushes with next-best-view planning~\cite{menon2023nbvsc,li2025enhancedviewplanning} to improve access through multiple layers of foliage, comparing dual-arm interaction with contact-tolerant single-arm planning~\cite{nemlekar2021lime}, and using incremental mapping~\cite{millane2024nvblox} to monitor canopy motion throughout the reach.

We believe that \textit{Push-and-Reach} opens new possibilities for agricultural robots to operate in dense canopies by actively modifying their surroundings through controlled physical interaction while maintaining collision-free access to the target.

\bibliographystyle{IEEEtran}
\bibliography{IEEEabrv,ref_final}

\appendix

\subsection{Planning and execution parameters}
\label{app:params}

\begin{table}[h!]
\centering
\footnotesize
\label{tab:parameters}
\setlength{\tabcolsep}{34pt}
\begin{tabular}{@{}ll@{}}
\toprule
Parameter & Value\\
\midrule
\multicolumn{2}{@{}l}{\emph{Trajectory optimization}}\\
Horizon $T$ & 30\\
Cost weights $(w_s,w_x,w_c)$ & $(1,10,5)$\\
Maximum joint step $\Delta_{\max}$ & 0.05~rad\\
Goal tolerances $(\varepsilon_p,\varepsilon_R)$ & $(1~\mathrm{cm},5^\circ)$\\
\midrule
\multicolumn{2}{@{}l}{\emph{Collision avoidance and pushing}}\\
Reach corridor radius $r_{\mathcal{R}}$ & 4~cm\\
Push stand-off $\delta_s$ & 2~cm\\
Maximum push length $\ell_{\max}$ & 12~cm\\
Arm separation $\delta_{\mathrm{arms}}$ & 5~cm\\
Reacher margin $s_r$ & 5~mm\\
Penetration bound $\delta$ & 5~cm\\
Pusher margin $s_p$ & $-\delta$\\
DBSCAN parameters $(\epsilon,m)$ & $(4~\mathrm{cm},5)$\\
\midrule
\multicolumn{2}{@{}l}{\emph{Execution}}\\
Control frequency & 100~Hz\\
Proportional gain $K_p$ & 2.0\\
Maximum push speed $v_{\max}^{\mathrm{push}}$ & 5~cm/s\\
\midrule
\multicolumn{2}{@{}l}{\emph{Bounds in Assumption~\ref{ass:err}}}\\
$(\varepsilon_c,\varepsilon_t,\varepsilon_d,\rho_B)$ & $(0.5,2.0,0.1,2.0)$~mm\\
Maximum body-point speed $v_B$ & 5~cm/s\\
Maximum sampling interval $h$ & 20~ms\\
\bottomrule
\vspace{-0.5cm}
\end{tabular}
\end{table}

\subsection{Robustness to normal estimation errors}
\label{app:normal-errors}

We evaluate the robustness of the push direction to perturbations of the contact normals shown in Fig.~\ref{fig:cluster}. Averaging the contact normals largely compensates for independent estimation errors, with a median push direction deviation of only $1.74^\circ$ for individual normal perturbations of $20^\circ$. Systematic errors have a greater effect on the push direction, as they affect all the normals in the same way. The selected contact cluster remains unchanged across the tested perturbations.

\end{document}